%% file: main.tex
\pdfoutput=1
\documentclass[11pt,letterpaper]{article}
\input{preamble}

\title{Robustness of breast lesion segmentation under MRI undersampling improves with k-space-aware deep learning}

\author{
    Lukas T.\ Rotkopf\textsuperscript{1,2,3}\thanks{Corresponding author: l.rotkopf@dkfz.de},
    Marco Schlimbach\textsuperscript{2},
    Julius C.\ Holzschuh\textsuperscript{3},\\
    Heinz-Peter Schlemmer\textsuperscript{3},
    Jens Kleesiek\textsuperscript{1,4,5,2,6,7},
    Moritz Rempe\textsuperscript{1,4}
    \vspace{2mm} \\
    \small \itshape
    \textsuperscript{1}Institute for AI in Medicine (IKIM), University Hospital Essen, Essen, Germany\\
    \small \itshape
    \textsuperscript{2}Department of Physics, Technical University Dortmund, Dortmund, Germany\\
    \small \itshape
    \textsuperscript{3}Division of Radiology, German Cancer Research Center (DKFZ), Heidelberg, Germany\\
    \small \itshape
    \textsuperscript{4}Cancer Research Center Cologne Essen (CCCE), University Medicine Essen, Essen, Germany\\
    \small \itshape
    \textsuperscript{5}RACOON Study Group, Site Essen, Essen, Germany\\
    \small \itshape
    \textsuperscript{6}German Cancer Consortium (DKTK), Partner Site Essen, Essen, Germany\\
    \small \itshape
    \textsuperscript{7}Medical Faculty and Faculty of Computer Science, University of Duisburg-Essen, Essen, Germany
}
\date{}

\begin{document}
\maketitle

\begin{abstract}
\noindent\textbf{Purpose:}
To assess whether breast lesion segmentation can be learned directly from acquired MRI k-space, and whether doing so improves robustness when data are accelerated or noisy.
\medskip

\noindent\textbf{Materials and Methods:}
This retrospective study used public breast dynamic contrast-enhanced MRI (DCE-MRI) datasets with acquired and synthetic k-space, together with a within-dataset synthetic control. We compared four 3D U-Net variants: a hybrid k-space-to-image model, a native k-space model, and magnitude and complex image-space baselines. Models were evaluated under increasing undersampling and added complex Gaussian k-space noise. The primary outcome was patient-level Dice similarity coefficient under cross-validation, with the hybrid model prespecified as the main comparison against the magnitude image-space baseline.
\medskip

\noindent\textbf{Results:}
At full sampling, the hybrid and image-space models performed similarly. As acceleration increased, the hybrid model retained substantially more segmentation accuracy and significantly outperformed the magnitude image-space baseline across moderate to high undersampling levels. The same pattern was observed when noise was added directly to k-space: the hybrid model degraded more slowly, whereas the image-space baseline failed under heavier noise. This advantage was reproduced in the within-dataset synthetic control. Feature analysis suggested that the k-space stage and image-space stage played complementary roles, with frequency-domain filtering concentrated before image-domain lesion localization.
\medskip

\noindent\textbf{Conclusion:}
K-space-aware deep learning improves the robustness of breast lesion segmentation under MRI undersampling and k-space noise, while matching image-space methods at full sampling.
\end{abstract}

\section*{Introduction}
\begin{figure*}[t]
\centering
\includegraphics[width=\textwidth]{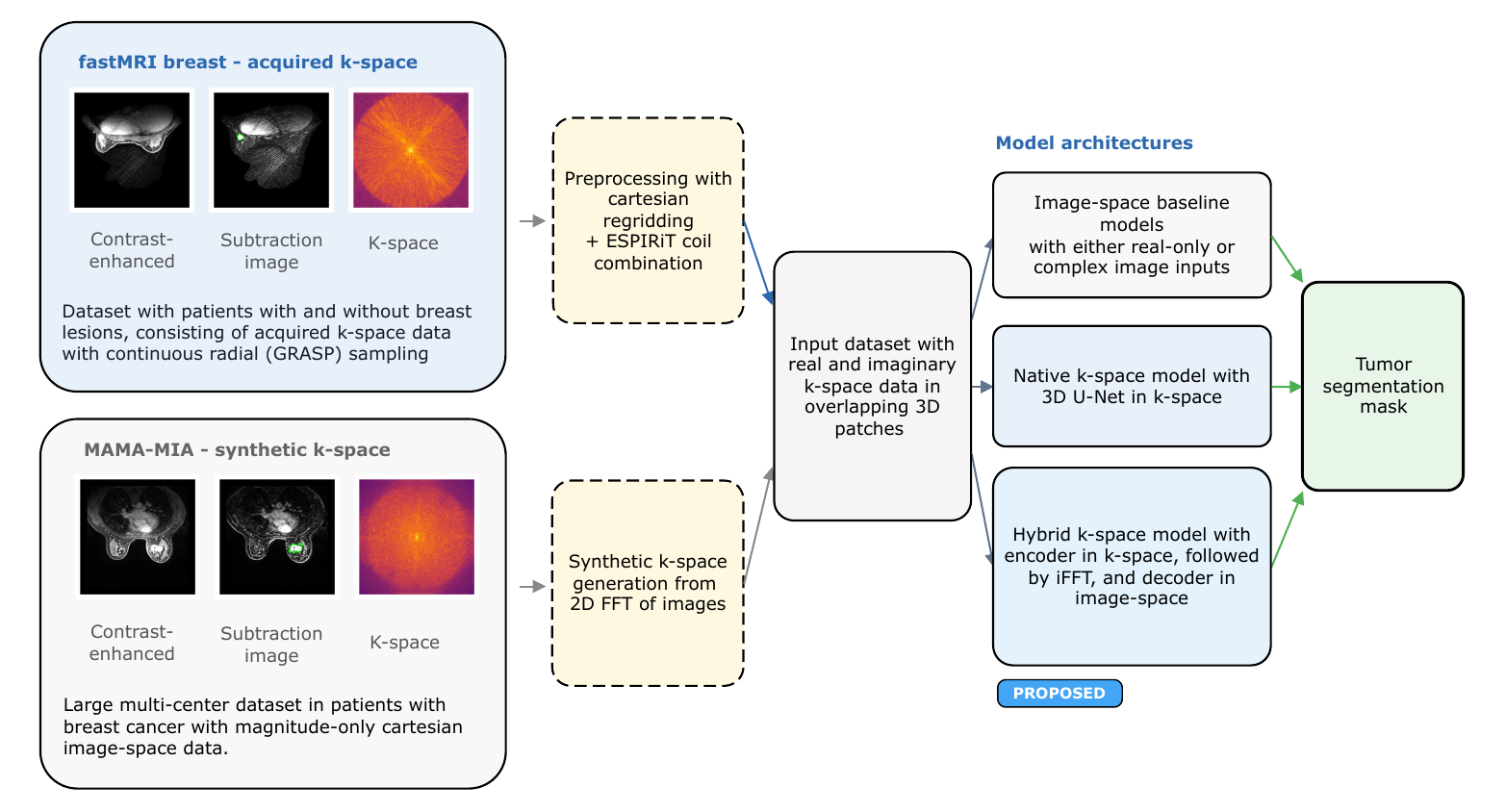}
\caption{Study overview. Two large-scale breast MRI datasets were used: fastMRI breast, providing acquired k-space data, and MAMA-MIA, providing synthetic k-space from reconstructed image-space data. To increase contrast agent conspicuity, temporal subtractions in k-space or image space between the pre-contrast and second post-contrast timepoints were used. Four different architectures were compared: the proposed hybrid k-space-to-image model, a native k-space model, and magnitude and complex image-space baseline models. DCE-MRI = dynamic contrast-enhanced MRI, FFT = fast Fourier transform, ESPIRiT = eigenvector-based SPIRiT}
\label{fig:overview}
\end{figure*}

Magnetic resonance imaging (MRI) is central to oncological diagnosis and treatment-response monitoring, but multi-sequence, multi-timepoint protocols are time-consuming and can limit access, throughput, and patient tolerance \citep{Kuhl2014}.

Undersampling shortens MRI by acquiring only part of k-space. Parallel imaging \citep{Griswold2002} and compressed sensing \citep{Lustig2007} routinely achieve fourfold to eightfold acceleration, and deep learning reconstruction has extended this range further \citep{Hammernik2018, Zbontar2018}. Most automated lesion segmentation is still performed after image reconstruction \citep{Hirsch2022}. Reconstruction can discard phase information \citep{Adams2017} and convert aggressive undersampling into aliasing artifacts before segmentation, so automated workflows may be constrained by reconstruction quality rather than acquired signal.

MRI data are acquired in k-space, a complex-valued frequency-domain representation. Low central frequencies encode coarse contrast, whereas peripheral high frequencies encode finer spatial detail. Before reconstruction, undersampling appears as missing samples; after reconstruction, those gaps become spatial aliasing. Direct k-space input can expose sampling patterns and signal components, including phase and acquisition-specific structure, that may be lost in image-space pipelines.

Direct k-space segmentation has been demonstrated for brain tissue classification \citep{Huang2019a, Kiefer2019}, skull stripping \citep{Rempe2024a}, cardiac structure delineation \citep{Schlemper2018, Zhang2024a, Li2025}, and knee joint segmentation \citep{Tolpadi2023, Desai2022}. However, prior work has largely used non-oncological tasks and synthetic k-space generated from reconstructed images \citep{Schlemper2018, Zhang2024a, Li2025, Huang2019a, Gosche2025}. Synthetic k-space has a clean, invertible relationship with image space, whereas acquired k-space contains noise correlations, trajectory imperfections, and residual coil-sensitivity patterns \citep{Pruessmann2006, Deshmane2012}. Magnitude-derived synthetic k-space also lacks phase. To our knowledge, acquired k-space has not been evaluated as direct input for oncological lesion segmentation.

We propose a hybrid k-space-to-image segmentation architecture with a learned k-space stage, fixed inverse Fourier transform bridge, and learned image-space stage. We apply it to breast dynamic contrast-enhanced MRI (DCE-MRI), a sensitive but time-consuming modality for breast cancer detection, screening, staging, and treatment-response assessment \citep{Mann2019}. Automated lesion segmentation is increasingly used for response assessment and quantitative downstream analysis, making breast DCE-MRI a relevant setting for testing whether direct k-space segmentation can support higher acceleration.

We hypothesized that a hybrid k-space-to-image architecture would be feasible for oncological lesion segmentation, competitive with image-space methods at full sampling, and more robust than image-space methods under undersampling. The purpose of this study was to evaluate this hypothesis using acquired k-space from the fastMRI breast dataset \citep{Solomon2025}, synthetic k-space from the MAMA-MIA dataset \citep{Garrucho2025}, and a within-dataset synthetic control derived from fastMRI breast reconstructions, as illustrated in Figure~\ref{fig:overview}.

\section*{Materials and Methods}
\subsection*{Study design}
This retrospective study used two publicly available large-scale breast DCE-MRI datasets. As both datasets were previously collected under institutional review board approval with appropriate consent or waiver at their originating institutions and were publicly released in de-identified form, no additional institutional review board approval was required. The study followed the Checklist for Artificial Intelligence in Medical Imaging (CLAIM) reporting guideline \citep{Mongan2020}.

\subsection*{Study cohorts and preprocessing}
The fastMRI breast dataset \citep{Solomon2025}, obtained from the NYU fastMRI Initiative database (fastmri.med.nyu.edu) \citep{Knoll2020,Zbontar2018}, provides acquired multi-coil radial k-space from golden-angle DCE-MRI examinations performed on a 3-T scanner (MAGNETOM TimTrio, Siemens Healthcare) between December 2019 and June 2022 at a single institution. The public release contains 284 patients (all female, mean age, 44 years $\pm$ 12) and 300 imaging studies with case-level labels indicating lesion status (negative, benign, or malignant). Patient-level Dice was reported for the subset of patients with confirmed malignancy for whom all visible lesions were manually delineated by a board-certified oncological radiologist with experience in breast imaging, resulting in 63 positive exams. Five-fold model training used the full available fastMRI breast cohort with stratification by lesion status.

The MAMA-MIA dataset \citep{Garrucho2025} is a multicenter collection of 1,506 pre-treatment breast DCE-MRI cases assembled from four collections in The Cancer Imaging Archive (TCIA). For the present analysis, all 1,506 cases with the provided expert annotations by breast cancer specialists were included.

\subsection*{Image and k-space preparation}
For fastMRI breast, the raw radial k-space was re-gridded onto a Cartesian grid and coil-combined into single-channel complex-valued k-space using ESPIRiT \citep{Uecker2014}. For MAMA-MIA, the synthetic k-space was generated by 2D Fourier-transforming the provided magnitude images. To increase lesion conspicuity, we subtracted the pre-contrast acquisition from the second post-contrast acquisition in the corresponding representation (k-space for fastMRI breast, image space for MAMA-MIA). Full preprocessing details are provided in Supplement S1.

During training, random undersampling was used as data augmentation to encourage a robust mapping from partial k-space rather than overfitting to specific aliasing patterns. Evaluation used the same fastMRI random-mask implementation with deterministic seeds derived from the acceleration factor, patient index, and patch position. Full acceleration and center-fraction configurations are listed in Supplement S2.

\begin{figure}[ht]
\centering
\includegraphics[width=0.6\textwidth]{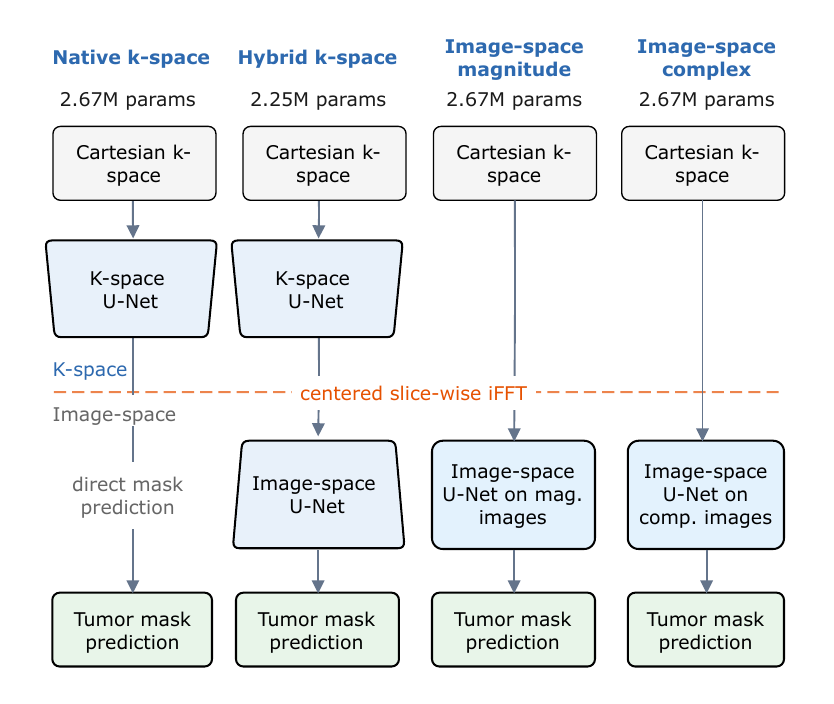}
\caption{Illustration of the model architectures compared in this study. FFT = fast Fourier transform.}
\label{fig:architectures}
\end{figure}

\subsection*{Models}

Four segmentation architectures sharing a common 3D U-Net backbone from MONAI were compared, as detailed in Supplement S3 \citep{MONAIConsortium2022}. All four architectures are illustrated in Figure~\ref{fig:architectures}.

The \textit{hybrid k-space-to-image model} chains a trainable k-space stage, a non-trainable inverse fast Fourier transform (FFT), and a trainable image-space stage. The k-space stage uses depth-only downsampling to preserve in-plane frequency resolution, whereas the image-space stage uses standard isotropic downsampling.

The \textit{native k-space model} is a 3D U-Net that operates entirely in k-space. For both training loss computation and inference, its outputs are converted to image-space class probabilities via a fixed inverse FFT.

The \textit{image-space magnitude baseline} is a conventional image-space 3D U-Net operating on magnitude images, obtained by slice-wise inverse FFT of k-space data, representing a conventional approach in which phase information is discarded.

The \textit{image-space complex baseline} is identical to the magnitude baseline except that it retains the real and imaginary components of the complex-valued image as two separate input channels to the network.

\subsection*{Training}
All models were trained using combined Dice and focal loss with per-class weighting; the native k-space model additionally used an auxiliary k-space mean squared error (MSE) term. Optimization used AdamW with cosine learning-rate scheduling, gradient clipping, and early stopping. Input volumes were processed as overlapping 3D patches with oversampling of lesion-containing patches. For the fastMRI breast dataset, additional positive-patient oversampling was used. Full training hyperparameters are provided in Supplement S2.

\subsection*{Outcomes and statistical analysis}
The primary outcome was the patient-level Dice similarity coefficient for the lesion class on lesion-positive fastMRI breast examinations. The primary comparison was the hybrid k-space-to-image model versus the image-space magnitude baseline across acceleration factors. Secondary analyses included the image-space complex baseline, the native k-space model, performance on synthetic k-space (MAMA-MIA), a within-dataset synthetic control using synthetic k-space derived from fastMRI breast reconstructions, catastrophic failure rate, and k-space noise robustness without undersampling.

Performance was evaluated using a stratified five-fold cross-validation with patient-level splits. fastMRI breast splits were stratified by lesion status and MAMA-MIA splits used shuffled patient-level folds because all included cases were lesion-positive. Results are reported as mean patient-level Dice with 95\% bootstrap confidence intervals. For fastMRI breast and the within-dataset synthetic control, the hybrid k-space-to-image model was compared with the image-space magnitude baseline at each acceleration factor using two-sided paired Wilcoxon signed-rank tests, with Holm correction across acceleration factors through 48$\times$ within each comparison family. The same paired Wilcoxon procedure with separate Holm correction was applied to the catastrophic failure rate and to the k-space noise robustness comparison. Holm-adjusted p $<$ 0.05 was considered statistically significant. Analyses were performed using Python 3.11 (Python Software Foundation, 2023), SciPy 1.16 (SciPy Developers, 2025), PyTorch 2.11 (PyTorch Foundation, 2026), and MONAI 1.5 (MONAI Consortium, 2025).

\begingroup
\small
\setlength{\tabcolsep}{4pt}
\begin{table*}[!ht]
\centering
\caption{Segmentation performance (mean Dice similarity coefficient) on acquired k-space (fastMRI breast) and synthetic k-space (MAMA-MIA) at selected acceleration factors.}
\label{tab:main_results}
\vspace{0.5em}
\begin{tabularx}{\textwidth}{l YYY YYY YYY}
\toprule
 & \multicolumn{3}{c}{\textbf{MAMA-MIA}} & \multicolumn{3}{c}{\textbf{fastMRI breast}} & \multicolumn{3}{c}{\textbf{fastMRI breast (synth.)}} \\
\cmidrule(lr){2-4} \cmidrule(lr){5-7} \cmidrule(lr){8-10}
 & 1$\times$ & 4$\times$ & 16$\times$ & 1$\times$ & 4$\times$ & 16$\times$ & 1$\times$ & 4$\times$ & 16$\times$ \\
\midrule
Image-space mag       & \dc{0.68} & \dc{0.63} & \dc{0.30} & \dc{0.52} & \dc{0.45} & \dc{0.07} & \dc{0.50} & \dc{0.45} & \dc{0.05} \\
Image-space complex   & \dc{0.68} & \dc{0.65} & \dc{0.34} & \dc{0.46} & \dc{0.39} & \dc{0.04} & \dc{0.47} & \dc{0.44} & \dc{0.07} \\
Native k-space        & \dc{0.54} & \dc{0.52} & \dc{0.31} & \dc{0.28} & \dc{0.25} & \dc{0.09} & \dc{0.30} & \dc{0.30} & \dc{0.13} \\
\midrule
Hybrid k-space        & \dc{0.67} & \dc{0.65} & \dc{0.38} & \dc{0.51} & \dc{0.46} & \dc{0.23} & \dc{0.55} & \dc{0.51} & \dc{0.28} \\
\bottomrule
\end{tabularx}

\vspace{0.3em}
\noindent Hybrid k-space-to-image = k-space stage followed by a fixed inverse FFT and an image-space stage; image-space magnitude = inverse fast Fourier transform (FFT) followed by magnitude extraction; image-space complex = inverse FFT with real and imaginary channels preserved; native k-space = 3D U-Net operating entirely in k-space.
\end{table*}
\endgroup

\section*{Results}
\subsection*{Model comparison on acquired k-space}

\begin{figure*}[!t]
\centering
\includegraphics[width=\textwidth]{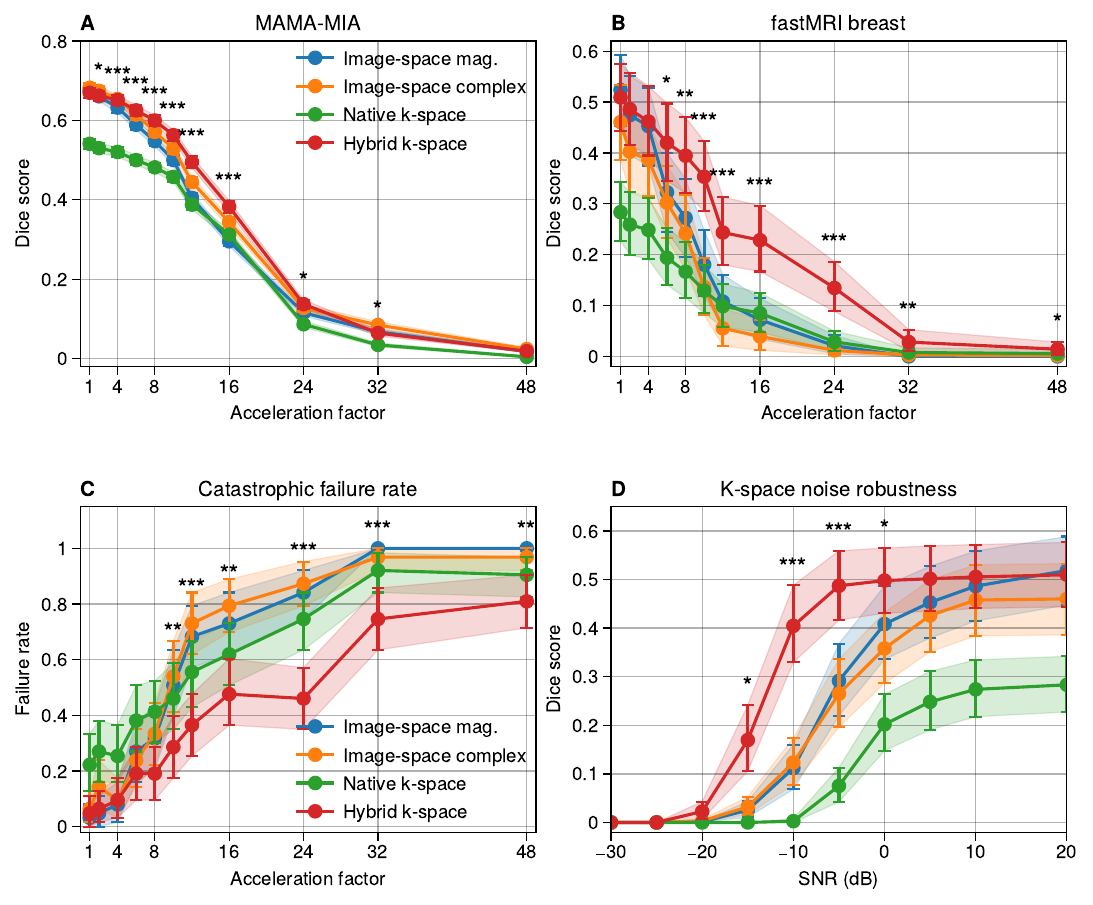}
\caption{Segmentation performance and robustness analyses. (A) MAMA-MIA dataset with synthetic k-space. (B) fastMRI breast dataset with acquired k-space. (C) Catastrophic failure rate, defined as patient-level Dice equal to zero, on fastMRI breast. (D) K-space noise robustness on fastMRI breast without undersampling. Shaded regions indicate 95\% bootstrap confidence intervals. Asterisks indicate Holm-adjusted significance for the hybrid k-space-to-image model versus the image-space magnitude baseline ($*$ p $<$ .05, $**$ p $<$ .01, $***$ p $<$ .001).}
\label{fig:dice_curves}
\end{figure*}

\begin{figure*}[!t]
\centering
\includegraphics[width=\textwidth]{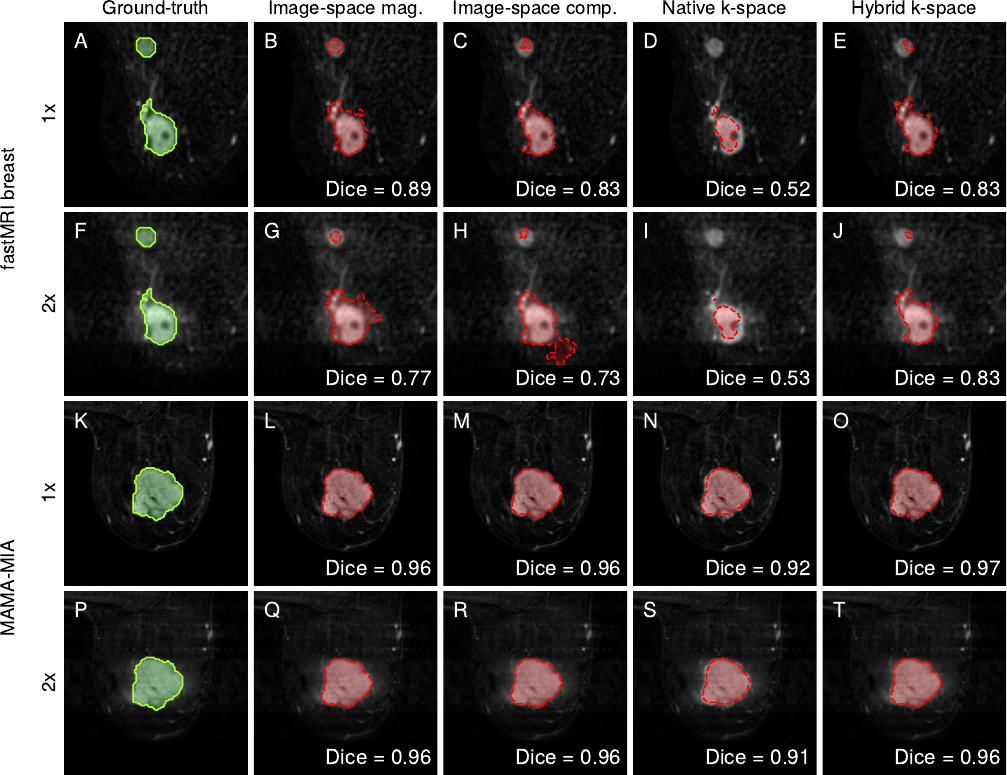}
\caption{Representative segmentation outputs at 1$\times$ and 2$\times$ acceleration. Top two rows: fastMRI breast. Bottom two rows: MAMA-MIA. Within each dataset, rows correspond to 1$\times$ and 2$\times$ acceleration. The first column shows the input image with the ground-truth mask (green); the remaining columns show predictions from each model with predicted lesion overlays (red). Per-example Dice coefficients are annotated.}
\label{fig:examples}
\end{figure*}

On fully sampled acquired k-space, the hybrid k-space-to-image model and the image-space magnitude baseline achieved comparable Dice coefficients, as shown in Table~\ref{tab:main_results} and Figure~\ref{fig:dice_curves}. As acceleration increased, performance declined for both architectures, but the decline was smaller for the hybrid model. The two models remained similar at fourfold acceleration and separated more clearly at higher acceleration. In the prespecified primary comparison, the hybrid model outperformed the image-space magnitude baseline from 6-fold through 48-fold acceleration after Holm correction (all adjusted p $<$ 0.05). At 16-fold acceleration, mean Dice was 0.23 for the hybrid model and 0.07 for the magnitude baseline. The image-space complex baseline was generally worse than the magnitude baseline, and the native k-space model performed substantially worse than the hybrid model at full sampling and moderate acceleration. Representative segmentation outputs are shown in Figure~\ref{fig:examples}.

Acceleration also increased the number of complete segmentation failures. For the image-space magnitude baseline, the catastrophic failure rate reached 100\% at 32-fold and 48-fold acceleration. The hybrid k-space-to-image model had lower failure rates from 10-fold through 48-fold acceleration (all Holm-adjusted p $<$ 0.01), with non-zero segmentations in most patients through 24-fold acceleration and in approximately 25\% of patients at 32-fold acceleration, as shown in Figure~\ref{fig:dice_curves}C. The image-space complex baseline failed more often than the magnitude baseline across most acceleration factors (e.g., 0.79 vs 0.73 at 16-fold acceleration). In the noise experiment, the hybrid model was less sensitive to added complex Gaussian k-space noise. At $-5$~dB and $-10$~dB signal-to-noise ratio (SNR), mean Dice was 0.49 and 0.41 for the hybrid model, compared with 0.29 and 0.11 for the magnitude baseline. Differences between the hybrid and magnitude models were significant from 0~dB down to $-15$~dB SNR (all Holm-adjusted p $<$ 0.05).

\subsection*{Effect of synthetic versus acquired k-space}

On synthetic k-space from MAMA-MIA, the hybrid k-space-to-image model and the image-space baselines achieved similar Dice coefficients at full sampling and remained closer across acceleration factors than on acquired data, as shown in Table~\ref{tab:main_results} and Figure~\ref{fig:dice_curves}. In the within-dataset synthetic control derived from fastMRI breast reconstructions, the hybrid model significantly outperformed the image-space magnitude baseline from 6-fold through 48-fold acceleration after Holm correction, as shown in Table~\ref{tab:main_results} and Figure~\ref{fig:dice_curves_syn}.

\begin{figure*}[!t]
\centering
\includegraphics[width=\textwidth]{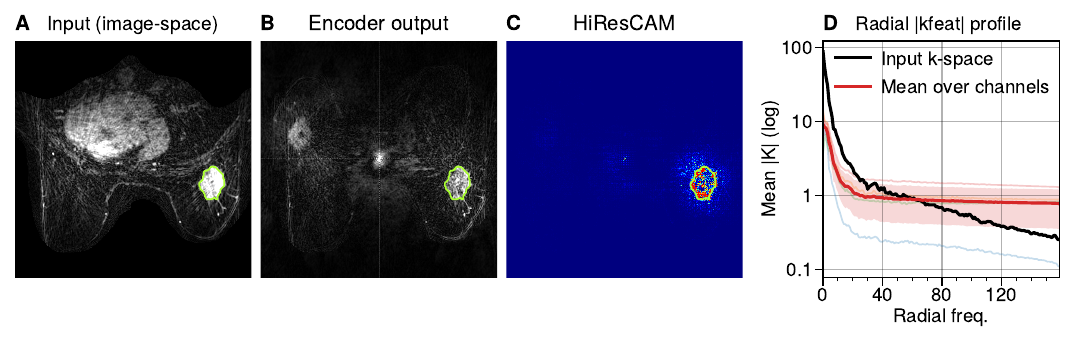}
\caption{Feature visualization of the hybrid k-space-to-image model's k-space stage on fully sampled fastMRI breast data. (A) Input image reconstructed via inverse FFT with overlaid lesion mask. (B) Phase-aligned coherent sum of the four output feature channels after transformation to image space. (C) Gradient-modulated coherent sum, weighting each feature channel by its gradient contribution to the lesion class. (D) Per-channel radial k-space energy profiles (colored traces) compared with the input k-space (black) and mean over channels (red, $\pm$1 SD shaded). FFT = fast Fourier transform.}
\label{fig:features}
\end{figure*}

\subsection*{Learned frequency-domain representations}
To characterize how the k-space stage contributed to segmentation, learned feature representations were visualized using phase-coherent summation and gradient-based attribution, as shown in Figure~\ref{fig:features}. In the phase-aligned coherent sum, k-space-stage outputs emphasized sharper structures and a more even distribution of frequency energy, consistent with learned frequency filtering. By contrast, gradient-weighted features localized most strongly around lesions after the image-space stage. These findings indicate that the two parts of the hybrid model contributed differently: the k-space stage modified frequency content before reconstruction, whereas the image-space stage provided spatially localized lesion segmentation.

\section*{Discussion}
In this retrospective study, a hybrid k-space-to-image architecture was more robust than image-space baselines when acquired breast DCE-MRI k-space was retrospectively accelerated. The image-space magnitude baseline failed in all patients at 32-fold acceleration and higher, whereas the hybrid model produced non-zero segmentations in most patients through 24-fold acceleration and roughly one quarter at 32-fold. The hybrid model also showed a significant advantage in the within-dataset synthetic fastMRI breast control from 6-fold through 48-fold acceleration; the MAMA-MIA synthetic k-space comparison was descriptive and showed smaller differences. These results suggest that k-space-aware models can use frequency-domain information relevant to segmentation, although acquisition-specific signal properties require prospective validation.

On acquired k-space, the performance gap widened as acceleration increased. K-space preserves information from signal reception, coil combination, and sampling that is not fully retained after reconstruction. At high acceleration, a k-space stage can act before missing samples become aliasing artifacts. The within-dataset synthetic control showed that the hybrid architecture can also improve robustness when k-space is generated from fastMRI breast reconstructions. Feature visualization supports this interpretation: the k-space stage performed spectral filtering and sharpening while preserving spatial structure, whereas lesion-specific activation arose mainly in the image-space stage. The noise experiment was consistent with this role, with learned frequency-domain preprocessing suppressing noise before image-domain segmentation.

The native k-space model provides complementary evidence. On synthetic data, it performed better than on real data, consistent with prior work showing that frequency-domain segmentation becomes more tractable when k-space and image space are concordant \citep{Li2024, Zhang2024a}. On acquired data, the native model performed poorly, likely reflecting the difficulty of localized feature extraction in the frequency domain when the Fourier relationship is complicated by acquisition imperfections. Its auxiliary k-space MSE term may also bias the model toward exact spectral reconstruction of the binary target rather than robust acquisition-domain features. The hybrid architecture avoids this by using a fixed inverse Fourier transform as a bridge to spatial segmentation.

The complex-valued image-space baseline did not improve over the magnitude baseline on real acquired k-space despite preserving phase information. On synthetic data, the complex baseline was comparable to the magnitude baseline. In MAMA-MIA, synthetic k-space was generated from magnitude-only reconstructions, so the imaginary input channel is nearly zero at full sampling and becomes informative mainly through undersampling artifacts. That comparison primarily reflects sensitivity to aliasing rather than phase. The weaker performance on acquired data may indicate that real phase patterns are more variable and harder to learn as additional image-space channels.

Absolute Dice values were moderate compared with published breast lesion segmentation on fully reconstructed multi-sequence data, where dedicated architectures often reach 0.7 to 0.85 \citep{Hirsch2022}. This likely reflects the limited size of the fastMRI breast lesion-positive subset and constraints on model complexity. The aim was not to maximize accuracy, but to compare architectures under matched conditions; relative trends were the primary outcome.

These findings have clinical implications for accelerated breast MRI. At higher acceleration, the hybrid model maintained measurable segmentation performance while image-space methods degraded substantially. This aligns with the Knee k-space to Segmentation (K2S) challenge, where reconstruction quality did not necessarily correlate with downstream segmentation accuracy \citep{Tolpadi2023}. K-space-aware processing could support more aggressive abbreviated protocols \citep{Kuhl2014}. The comparison across acquired and synthetic k-space also extends concerns that synthetic k-space can inflate reconstruction performance \citep{Shimron2022} to segmentation tasks.

This study had limitations. Both datasets came from limited clinical settings, and generalization to other anatomies, field strengths, and scanner platforms remains to be established. The fastMRI breast lesion-positive subset was small. The hybrid model also requires approximately 9-fold higher computational cost than image-space baselines, although lightweight encoders or distillation could reduce this overhead. Undersampling was applied retrospectively with Cartesian masks after regridding rather than on the native radial trajectory. Prospective validation on accelerated acquisitions is needed.

In conclusion, a hybrid k-space-to-image architecture showed increasing advantage over image-space segmentation approaches at higher acceleration on acquired breast DCE-MRI k-space and in a within-dataset synthetic fastMRI breast control. Prospective validation on accelerated acquisitions is warranted.

\section*{Data availability}
Both datasets used in this study are publicly available, fastMRI breast at \href{https://github.com/TIO-IKIM/kspace-breast-seg}{https://fastmri.med.nyu.edu}, and MAMA-MIA at \href{https://github.com/TIO-IKIM/kspace-breast-seg}{https://www.cancerimagingarchive.net}. Processing and analysis code is available at \href{https://github.com/TIO-IKIM/kspace-breast-seg}{https://github.com/TIO-IKIM/kspace-breast-seg}.

\section*{Acknowledgements}
This study was supported by the J\"oster Foundation.

\section*{Competing interests}
The authors have no competing interests to declare.

{\small
\bibliographystyle{ieeetr}
\bibliography{references}
}

\appendix
\renewcommand{\thefigure}{S\arabic{figure}}
\renewcommand{\thetable}{S\arabic{table}}
\setcounter{figure}{0}
\setcounter{table}{0}

\section*{Supplement S1. Dataset preprocessing}
\label{supp:preprocessing}

\subsection*{fastMRI breast}
The fastMRI breast dataset provides raw multi-coil radial k-space from contrast-enhanced golden-angle (GRASP) DCE breast MRI \citep{Solomon2025}. Raw data were preprocessed as follows. Partition k-space was zero-filled and Fourier-transformed along the slice-encoding direction to obtain individual slices. Density-compensated non-uniform fast Fourier transform (NUFFT) regridding was then performed for each slice onto a $320 \times 320$ Cartesian grid. ESPIRiT coil sensitivity estimation was computed from a $32 \times 32$ calibration region, and coil combination with ESPIRiT resulted in single-channel complex-valued Cartesian k-space. Per-patient intensity normalization was performed by dividing all time points by the root-mean-square (RMS) magnitude of the pre-contrast k-space. The primary network input was the complex-valued temporal subtraction k-space (second post-contrast minus pre-contrast to maximize contrast enhancement), split into real and imaginary channels. During training, additional pseudo-volumes from the first post-contrast minus the pre-contrast timepoints were included as augmentation, reusing the same target and segmentation mask.

\subsection*{Synthetic fastMRI breast control}
To isolate acquisition-derived information from architectural differences, a within-dataset synthetic k-space variant of fastMRI breast was generated. The dataset-provided total variation (TV)-regularized golden-angle reconstructions were 2D Fourier-transformed per slice with orthonormal normalization to produce synthetic complex-valued k-space. All downstream processing (per-patient pre-contrast RMS scaling, second-minus-pre-contrast subtraction in k-space, augmentation timepoint pairs, and mask handling) followed the same pipeline as the acquired-k-space variant.

\subsection*{MAMA-MIA}
The MAMA-MIA dataset provides reconstructed DICOM images from four TCIA collections \citep{Garrucho2025}. Images were resampled to $1$~mm isotropic resolution using B-spline interpolation for images and nearest-neighbor for segmentation masks. Temporal subtraction images (second post-contrast minus pre-contrast) were computed in image space, padded or center-cropped to $384 \times 384$ pixels, and multiplied by a cosine-taper window in padded boundary regions to reduce Gibbs ringing artifacts. Synthetic complex-valued k-space was then generated by applying a two-dimensional centered FFT with orthonormal normalization.

\section*{Supplement S2. Training and evaluation details}
\label{supp:training}
\subsection*{Undersampling configurations}
The acceleration factors and corresponding center fractions used for evaluation were: 1x (fully sampled), 2x (center fraction 0.04), 4x (0.08), 6x (0.05), 8x (0.04), 10x (0.03), 12x (0.02), 16x (0.015), 24x (0.008), 32x (0.004), and 48x (0.002) on both datasets.

\subsection*{Training hyperparameters}
Input volumes were processed as 3D patches of 24 slices with a stride of 16 slices along the depth dimension. Spatial dimensions were, as stated above, $320 \times 320$ for fastMRI breast and $384 \times 384$ for MAMA-MIA. Lesion-containing patches were oversampled to a target positive fraction of 50\%; fastMRI breast also used positive-patient oversampling with factor 2. During training, random Cartesian undersampling was applied as data augmentation with probability 0.3, drawing from twofold acceleration with center fraction 0.04 and fourfold acceleration with center fraction 0.08. Training ran for a maximum of 60 epochs with early stopping monitoring exam-level validation Dice.

The combined loss used Dice weight 0.7, focal weight 0.3, focal $\gamma = 1.5$, and per-class weights of 0.05 (background) and 0.95 (lesion). The auxiliary k-space MSE loss for the native k-space model used weight 0.5. Optimization used AdamW (learning rate $3 \times 10^{-4}$) with cosine annealing to $\eta_{\min}=1 \times 10^{-6}$ over 60 epochs, mixed precision, and gradient clipping at norm 1.0. Training batch size was 16 on fastMRI breast and 24 on MAMA-MIA.

\subsection*{Evaluation details}
Five-fold cross-validation used patient-level assignment with seed 123. fastMRI breast used stratified folds, whereas MAMA-MIA used shuffled K-fold assignment because all included cases were lesion-positive. For inference, overlapping patch predictions (depth 24, stride 16) were combined by strict majority voting, whereas binary foreground predictions were summed across overlapping patches, and a voxel was assigned to the foreground class only when the sum exceeded half the overlap count. The reported Dice was computed per patient on the reconstructed full volume, with the fastMRI breast Dice being averaged over lesion-positive patients only. Evaluation undersampling masks used deterministic seeds derived from the acceleration factor, patient index, and patch position. K-space noise robustness was evaluated without undersampling by adding complex Gaussian noise at SNR values of 20, 10, 5, 0, -5, -10, -15, -20, -25, and -30 dB.

\begin{figure*}[ht]
\centering
\includegraphics[width=\textwidth]{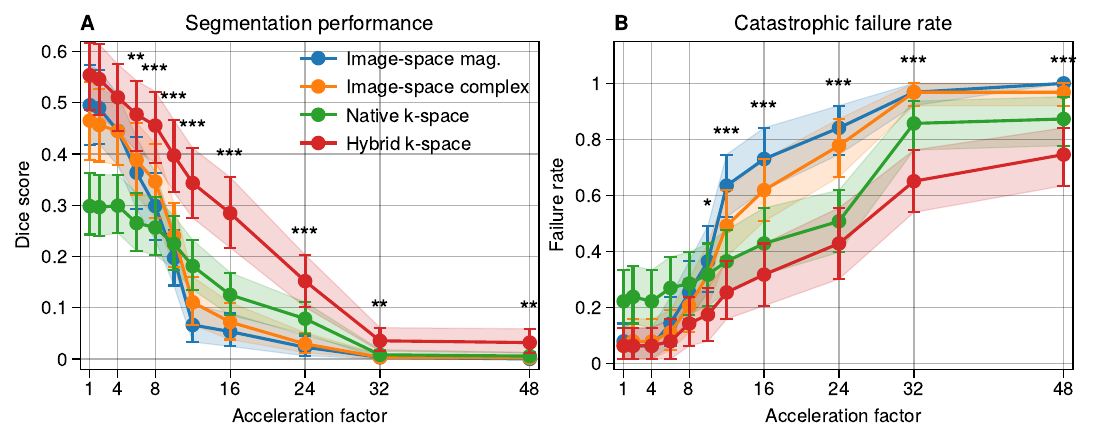}
\caption{Segmentation performance (Dice similarity coefficient) and catastrophic failure rate on the within-dataset synthetic fastMRI breast control. Shaded regions indicate 95\% bootstrap confidence intervals. Asterisks denote Holm-adjusted $P < .05$ for the hybrid k-space-to-image model versus the image-space magnitude baseline.}
\label{fig:dice_curves_syn}
\end{figure*}

\subsection*{Software and hardware}
All analyses used Python 3.11, SciPy 1.16, PyTorch 2.11, PyTorch Lightning 2.6, MONAI 1.5, and SigPy 0.1. Experiments were conducted on an NVIDIA GH200 Grace Hopper server (96~GB HBM3 GPU memory, 72-core ARM Neoverse-V2 CPU, 576~GB system memory).

\section*{Supplement S3. Architecture details}
\label{supp:architectures}

\begin{table}[htp]
\centering
\caption{Model architecture and computational complexity. MACs and inference time measured on a single GPU.}
\label{tab:complexity_supp}
\setlength{\tabcolsep}{2pt}
\begin{tabular}{l c c c}
\toprule
Model & Params (M) & MACs (G) & Time (ms) \\
\midrule
Native k-space                & 2.67 & 32.6 & 12.5 \\
Image-space magnitude         & 2.67 & 32.2 & 12.6 \\
Image-space complex           & 2.67 & 32.6 & 12.6 \\
Hybrid k-space-to-image       & 2.25 & 279.7 & 132.2 \\
\bottomrule
\end{tabular}

\vspace{0.3em}
\small
\noindent M = million; G = billion; ms = milliseconds; MACs = multiply-accumulate operations.
\end{table}

All architectures used the MONAI 3D U-Net implementation. Real and imaginary components were treated as separate channels. Convolutional blocks used kernel size 3 unless stated otherwise, 2 residual units per level, group normalization (2 groups), leaky ReLU activations with negative slope 0.1, and no dropout. The image-space U-Net used 3 levels in all architectures; the hybrid k-space-to-image model's k-space stage used 2 levels.

The image-space magnitude baseline and image-space complex baseline used the same 3D U-Net backbone with base channel count 24. Both first applied a fixed inverse FFT to the complex k-space input. The magnitude baseline then supplied only the magnitude image to the network, whereas the complex baseline retained real and imaginary image channels. The native k-space model also used base channel count 24 and operated entirely in k-space, predicting one complex-valued k-space per class that was converted to per-class image-space probabilities via a fixed inverse FFT followed by magnitude and a calibration head.

The reported hybrid k-space-to-image architecture used hidden factor 20 and four complex-valued intermediate feature channels in the k-space stage, followed by a fixed inverse FFT bridge and a learned image-space stage. In the k-space stage, depth-only downsampling with a stride of 2 along the depth dimension and 1 along the spatial dimensions, and upsampling kernels of $3 \times 1 \times 1$, preserved in-plane frequency resolution. The image-space stage used isotropic stride-2 downsampling with $3 \times 3 \times 3$ upsampling kernels.

The hybrid k-space-to-image model contained fewer trainable parameters than the baselines but required substantially more computation due to the lack of spatial downsampling in the k-space stage, as shown in Table~\ref{tab:complexity_supp}.

\end{document}

%% file: preamble.tex
\PassOptionsToPackage{table}{xcolor}

\usepackage[letterpaper,margin=1in]{geometry}
\usepackage[T1]{fontenc}
\usepackage[utf8]{inputenc}
\usepackage{lmodern}
\usepackage{microtype}
\usepackage{graphicx}
\usepackage{amsmath,amssymb}
\usepackage{booktabs}
\usepackage{pgf}
\usepackage{tabularx}
\usepackage{array}
\newcolumntype{Y}{>{\centering\arraybackslash}X}

\usepackage[round,numbers,sort&compress]{natbib}

\usepackage{xcolor}
\usepackage{colortbl}
\definecolor{linkblue}{rgb}{0.21,0.49,0.74}
\usepackage[breaklinks,colorlinks,allcolors=linkblue]{hyperref}

\makeatletter
\newcommand{\dc@min}{0}
\newcommand{\dc@max}{0.75}
\newcommand{\dcsetup}[2]{%
  \renewcommand{\dc@min}{#1}%
  \renewcommand{\dc@max}{#2}%
}
\newcommand{\dc@color}[1]{%
  \pgfmathparse{min(1, max(0, (#1 - \dc@min)/(\dc@max - \dc@min)))}%
  \xdef\dc@ratio{\pgfmathresult}%
  \ifdim\dc@ratio pt > 0.5pt%
    \pgfmathparse{(\dc@ratio - 0.5)*2}%
    \xdef\dc@t{\pgfmathresult}%
    \pgfmathparse{round(255 + \dc@t*(60 - 255))}%
    \xdef\dc@r{\pgfmathresult}%
    \pgfmathparse{round(235 + \dc@t*(180 - 235))}%
    \xdef\dc@g{\pgfmathresult}%
    \pgfmathparse{round(120 + \dc@t*(75 - 120))}%
    \xdef\dc@b{\pgfmathresult}%
  \else%
    \pgfmathparse{\dc@ratio * 2}%
    \xdef\dc@t{\pgfmathresult}%
    \pgfmathparse{round(230 + \dc@t*(255 - 230))}%
    \xdef\dc@r{\pgfmathresult}%
    \pgfmathparse{round(70 + \dc@t*(235 - 70))}%
    \xdef\dc@g{\pgfmathresult}%
    \pgfmathparse{round(50 + \dc@t*(120 - 50))}%
    \xdef\dc@b{\pgfmathresult}%
  \fi%
  \cellcolor[RGB]{\dc@r,\dc@g,\dc@b}%
}
\newcommand{\dc}[1]{\leavevmode\dc@color{#1}#1}
\makeatother

\newcommand{\TODO}[1]{}